\documentclass{IOS-Book-Article}
\usepackage{lipsum}
\usepackage{mathptmx}
\usepackage{graphicx}
\usepackage{multicol}
\usepackage{xcolor}
\usepackage{tikz}
\usepackage{amsfonts} 
\usepackage{tabularx}
\newcommand{\para}[1]{\paragraph{\textnormal{\textbf{#1}}.}}

\usepackage[show]{chato-notes}

\newcommand{\tikzxmark}{%
\tikz[scale=0.23] {
    \draw[line width=0.7,line cap=round] (0,0) to [bend left=6] (1,1);
    \draw[line width=0.7,line cap=round] (0.2,0.95) to [bend right=3] (0.8,0.05);
}}

\usepackage{soul}\setuldepth{article}
\def\hb{\hbox to 11.5 cm{}}

\begin{document}

\pagestyle{headings}
\def\thepage{}
\begin{frontmatter}              

\title{Annotating Topical Legal Insights from Case Proceedings}

\markboth{}{November 2023\hb}

 \author[A]{\fnms{Subinay} \snm{Adhikary}
},
\author[A]{\fnms{Dwaipayan} \snm{Roy}
},
\author[B]{\fnms{Debasis} \snm{Ganguly}
},
\author[C]{\fnms{Shouvik} \snm{Kumar Guha}
}
and
\author[A]{\fnms{Kripabandhu} \snm{Ghosh}
}

\address[A]{Indian Institute of Science Education and Research Kolkata, India}
\address[B]{University of Glasgow, United Kingdom}
\address[C]{West Bengal National University of Juridical Sciences, India}

\begin{abstract}
In this paper, we mainly concentrate on finding concepts or topics from the legal case proceedings, since adopting a structured representation for legal documents, as opposed to a mere bag-of-words flat text representation, can significantly enhance processing capabilities. To achieve this objective, we put forward a set of diverse concepts for legal case proceedings. With this motivation, we propose LeDA, a system for \textbf{Le}gal \textbf{D}ata \textbf{A}nnotation. The system offers the generic functionality of annotating and adjudicating entities or concepts within documents via a web-based interface. A novel feature of our system is that it allows to dynamic create new tags for annotation, which is a particularly useful provision for situations where there exists no pre-defined ontology for the entities (concepts) that need to be annotated - these being rather discovered by annotators as they continue examining more documents. The system that we demonstrate is currently in use to annotate a set of concepts from legal documents to construct semantic representations of documents as bags of concepts that can then be used for several downstream tasks, such as prior case retrieval, judgment prediction, and so on. Along with the system features in general, we also describe how LeDA was used by 3 assessors to annotate and adjudicate legal concept names from Indian Supreme Court case proceedings. 
\end{abstract}

\end{frontmatter}
\markboth{November 2023\hb}{November 2023\hb}
 \section{Introduction}
 Legal documents are considerably large in size and complex in nature~\cite{Shukla2022}, making \textit{information extraction} from legal documents is a challenging task for the research community. Legal case documents consist of several pieces of information such as who are parties to the case (\textit{appellant} and \textit{respondent}), in which court (e.g. \textit{Supreme court, High court, District court}) the case is appealed, the motive of the incident, the judgment of the case, etc.
Some techniques (e.g. catch phrase extraction~\cite{r1,r2}, evidence extraction~\cite{r4, adhikary2023automated} and witness testimony extraction\cite{r3} etc.) are introduced by the research community to extract aforementioned information from legal documents. Catchphrases are noun phrases that have been extracted from the document and were selected based on term frequency-inverse
\par\rule{0.9\textwidth}{1 pt} 
 \par
 \noindent This paper is an extended version of our JURIX 2023 Demo paper, \textit{LeDA: A System for Legal Data Annotation}, accepted at the \href{https://cris.maastrichtuniversity.nl/en/activities/workshop-on-annotation-of-legal-data-hosted-by-jurix-2023/}{\textit{Annotation of Legal Data}} Workshop. \\
   term frequency score. On the other hand, to identify sentences of witness testimony (e.g. \texttt{The body of Gian Kaur was sent to Dr. Singh (PW 6) for post-mortem who noticed five minor injuries on the body of the deceased}), the authors have leveraged NLP techniques like linguistic knowledge-base and distant supervision (e.g. Bi-LSTM \cite{r5}). Although these methods are useful for searching information from documents, none of them are capable of gaining a \textit{topical view or thematic view} from the documents. 
\begin{figure}[t]
    \centering
     \caption{Blue colored words are the example of \textit{Catchphrases}, where {\it `murder'}, {\it `life imprisonment'}, and {\it `parole'} provide some information in this span of text. Although unable to describe the \textit{thematic view} (i.e., \textbf{a person murdered his wife during parole}).}
    \includegraphics[width=0.9\textwidth]{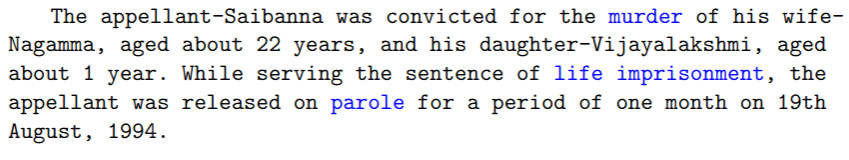}
   
    \label{fig:img2}
\end{figure}

 \noindent In Figure~\ref{fig:img2}, we highlighted words like `murder', `life imprisonment', and `parole' as phrases that can be extracted using different phrase extraction tools (e.g. KEA \cite{r6} ). However, the aforementioned paragraph represents an \textit{topic} or \textit{fine-grained information} (the term \textit{topic} and \textit{fine-grained information} will be used interchangeably in this paper) that is \textit{accused person murdered his wife during his parole}, so the event is \textit{Murder\_on\_parole} (while serving his parole, a man killed someone). To acquire this type of information, a legal practitioner needs to read the whole document, which is time-consuming and laborious. \\
However, to get the thematic view of the document, we need to extract fine-grained level information or the events from the document. Our primary objective is to find this kind of fine-grained level information from the document and label that information with suitable tags. The objective of our proposed annotation tool, called LeDA, is to reduce the effort of annotation of legal documents with such thematic concepts that effectively capture the \textbf{``aboutness"} of a case document.
\\

\begin{table}[t]
\centering
\caption{A set of tags and their descriptions used in LeDA.}
\scalebox{0.90}{%
\begin{tabularx}{\textwidth}{|l X |} 
 \hline 
 Legal concept category & Description\\ [0.5ex] 
 \hline\hline
 \multicolumn{2}{|c|}{Static Initialisation} \\
 \hline

\textit{
Murder\_on\_parole 
} &  murder during parole \\
\textit{
    Second\_murder
} &  committed second murder   \\

\textit{
    Homicide\_murder
} & 
    homicide amounting to murder
 \\

\textit{
Revenge
} & 
    Court identified as revenge
 \\
\textit{
    Property\_dispute 
}& 
committed as a result of property 
\\
\textit{
    Evidence\_inconsistency
} & 
    evidence of crime was not found 
\\
\textit{
    Evidence\_insufficient
} & 
    having been found inconclusive/insufficient
  \\

\textit{
    Testimony\_challenged
} & 
    witness testimony presented in favour of the prosecution or the defense
 \\ [1ex] 
 \hline
 \multicolumn{2}{|c|}{Dynamically added by legal experts during annotation} \\
 \hline

\textit{Investigation\_agency} & Type of cases were investigated by any Central institute/state institute (e.g., CBI, NIA, ED, CID). \\

\textit{Testimony\_Challenged} & This will reflect whether the witness testimony presented in favour of the prosecution or the defence has been contested by the other party and also whether the court has agreed to such challenge.\\ 
\hline
\end{tabularx}}

\label{tab:tag_list}
\end{table}

\section{Key features of LeDA}
In a standard sequence labeling annotation workflow, the task involves selecting spans of text, like entities and relations, from a document and categorizing them into predefined types. However, legal document annotation presents a unique challenge because the concepts to be annotated are not straightforward and atomic, like entity names. This complexity makes it difficult to rely on a static set of categories for annotating these concept types. Initially, when we attempted to use a conventional sequence labeling tool for annotation, we quickly realized the need for a more flexible solution that would allow annotators to create new concept types. This ability to create new tags is a central and innovative feature of LeDA.

Table~\ref{tab:tag_list} presents the predefined concept types, which were established in extensive consultations with legal experts, including criminal lawyers (from West Bengal National University of Juridical Sciences), along with the new tags that were introduced during the annotation process.

Another significant feature of our tool, particularly crucial in the realm of legal concept annotation, is the process of meta-annotation. This involves adjudicating multiple annotations carried out by different annotators, similar to a version control system's merge operation. By focusing on independent annotation, we aim to reduce biases, especially since shared documents may introduce bias. In essence, meta-annotators resolve conflicts by examining differing annotations and selecting one or none of the conflicting entries. LeDA provides a dual view of two distinct annotations of the same document, empowering a \textbf{meta-annotator} to reconcile differences. A comparison of LeDA with other annotation tools is outlined in Table~\ref{tab:Features}. Our code is publicly available on GitHub.\footnote{\url{https://github.com/subinayadhikary/LeDA}} \\
\begin{table}[t]
    \caption{Feature-wise comparison between different tools.}
    \resizebox{0.80\columnwidth}{!}
    {%
    
\begin{tabular}{ |c|c|c|c|c|c| } 
 \hline
 Feature & BRAT\footnote{\url{https://brat.nlplab.org/}} & GATE\footnote{\url{https://gate.ac.uk/}} & Label Studio\footnote{\url{https://labelstud.io/}} & UBIAI\footnote{\url{https://ubiai.tools/}}  & \textbf{LeDA} \\ 
 \hline
 Multiple tag & \tikzxmark & \tikzxmark & \tikzxmark & \checkmark &\checkmark  \\
 \hline
 Dynamic tag &\tikzxmark & \checkmark &\checkmark & \checkmark &\checkmark \\
 \hline 
 Adjudication & \tikzxmark & \tikzxmark & \tikzxmark & \checkmark & \checkmark\\
 \hline
 Highlight & \checkmark & \checkmark  & \checkmark & \checkmark & \checkmark \\
 \hline
 IAA calculation & \tikzxmark &\checkmark&\tikzxmark  & \tikzxmark  & \checkmark\\
 \hline
 Remote access & \tikzxmark & \tikzxmark & \tikzxmark & \checkmark & \checkmark\\
 \hline
 Cost  &Free  & Free  &Free  & Proprietary & Free\\
 \hline
 \end{tabular}}
    
    \label{tab:Features}
\end{table}
There are several existing tools such as BRAT, GATE, Doccano, YEDDA and DoTAT \cite{lin2022dotat}  available for general text annotation. However, when it comes to annotating legal data, some critical features, including the capability for handling multiple tags, calculating Inter-Annotator Agreement (IAA), and providing remote access, are notably absent in these tools. A comparison of these tools with LeDA, highlighting their available features, is presented in Table~\ref{tab:Features}. To assess the annotation process with input from actual legal experts and to evaluate the utility of LeDA's other features, we employed case judgments from the Indian Supreme Court\footnote{\url{https://indiankanoon.org/}}. Legal practitioners affiliated with the West Bengal National University of Juridical Sciences annotated 200 legal documents using LeDA. The feedback we received on LeDA's features was highly satisfactory, and significantly, no new feature suggestions were made, affirming the tool's effectiveness

\section{Annotation of the documents}\label{sec:overview}
The overall system consists of a frontend and a backend. The frontend 
is created by using HTML, CSS, and Javascript. In the backend, we use the python-based web framework Django. For hosting our annotation tool we use PythonAnywhere\footnote{\url{https://www.pythonanywhere.com/}} server. LeDA provides different interfaces for annotators and the super annotator.
\begin{figure*}[t]
    \centering
     \caption{LeDA workflow: `A': upload documents; `B': select a document from a list; `C' indicates that the document is annotated by both the annotators; `D' indicates the IAA score; `E': computes the IAA score; `F': button to delete a document; `G': button to add new a tag; `H': selected document; `I' set of tags; `J': search documents tag-wise; `K': buttons to add, remove or save the highlighted span and labels; `L': highlighted span; `M': label for highlighted span; `N': search a document.}
    \includegraphics[width=.90\textwidth]{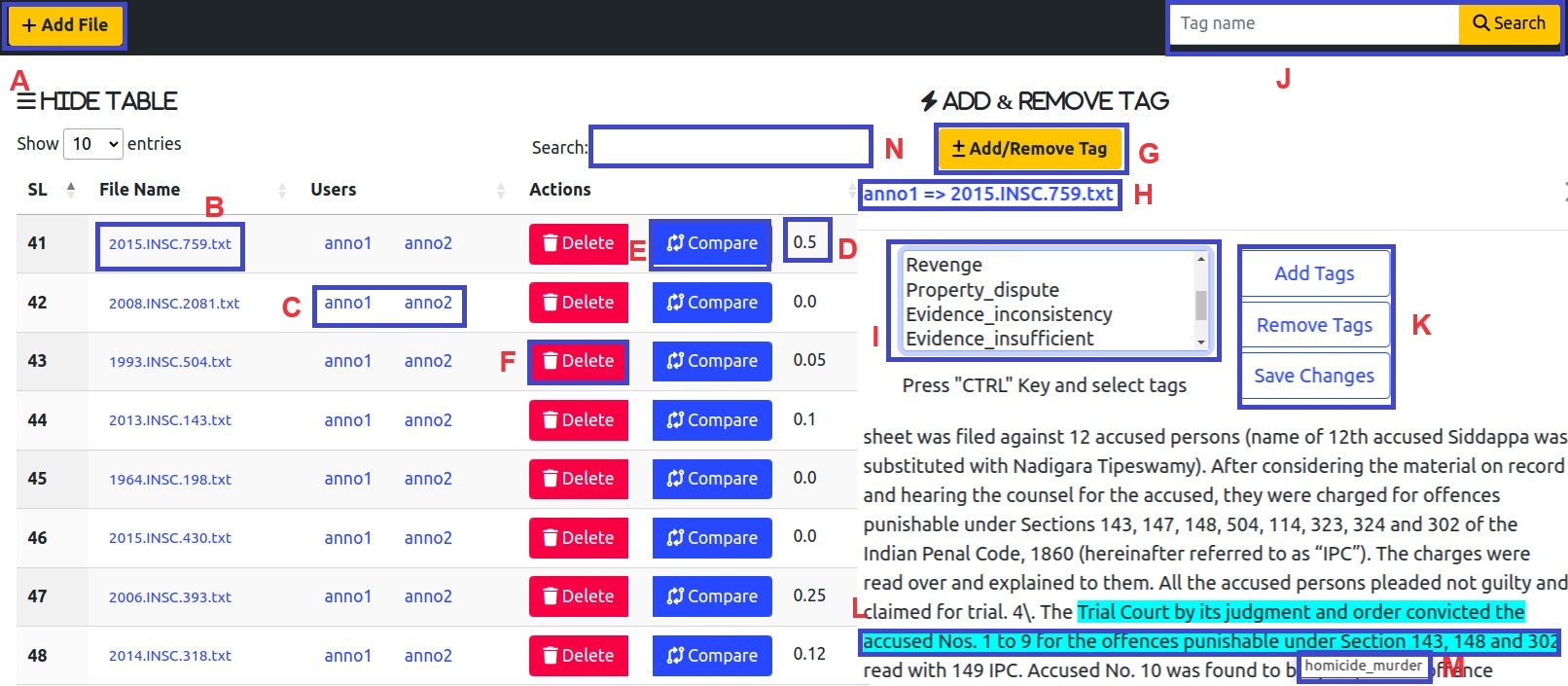}
   
    \label{fig:img1}
\end{figure*}

\para{Annotator view}
Every annotator is provided with a unique login ID and password, which are assigned by the administrator. Annotators utilize these credentials to access the interface, as illustrated in Figure~\ref{fig:img1}.
Annotators begin by selecting documents they have permission to annotate. They meticulously identify and categorize fine-grained data by associating tags from a predefined list with relevant words, using the `Add tags' function. This process entails highlighting and tagging specific details within the document. Once the annotation is complete, they can save the annotated data in JSON format by clicking the `Save changes' button. In cases where adjustments are necessary, the 'Remove tag' function enables the removal of specific tag-word links, offering flexibility for detailed annotation modifications. This iterative process is applied to various sets of words, allowing annotators to make comprehensive changes to their annotations. For example, as depicted in Figure~\ref{fig:img1}, the annotator's workflow involves the following steps:
\begin{itemize}
    \item Selecting a document (represented as `B').
    \item Highlighting a specific set of words (illustrated as `L' and `M').
    \item Associating the appropriate tags.
    \item Preserve these alterations by clicking the `Save Changes' button (depicted as `K') to update the JSON file.
\end{itemize}
Furthermore, annotators can use tags (referred to as `J' in Figure~\ref{fig:img1}) for searching and retrieving documents. The annotation process is initiated using a predefined list of tags. In cases where annotators encounter detailed information not covered by the current tag list, they have the option to request the super annotator to incorporate that specific fine-grained information into the existing set of tags.
\begin{figure}[t]
\center
    \caption{An instance where a new tag, specifically \texttt{Testimony\_challenged}, emerged during the annotation process due to a highlighted text span not aligning thematically with the predefined list of concept types (as shown in Table \ref{tab:tag_list})}.
    \includegraphics[width=.80\textwidth]{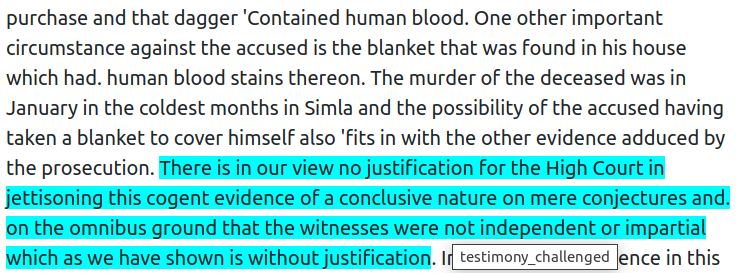}
    
    \label{fig:new_tag}
\end{figure}
\para{Super-annotator view}
Super annotator plays a crucial role after the first phase of annotation is complete, with greater privileges than annotators. As depicted in Figure~\ref{fig:img1}, super annotators possess the capabilities to upload, remove documents, initiate annotations, introduce new tags, and compute Inter-Annotator Agreement (IAA) \cite{DBLP:conf/jurix/BhattacharyaPG019}. Our approach to IAA computation represents a novel method, distinct from the established technique employed in GATE. Moreover, annotators have the option to request super annotators to add new tags to the existing list. Utilizing the `Add New Tag' function (as described in `G'), they augment the tag list, as illustrated in Figure~\ref{fig:new_tag}—reflecting the dynamic tag feature, which accommodates situations where annotators commence the annotation process without a predefined ontology. To quantify the quality of annotation, computation of the Inter-Annotator Agreement (IAA) plays a crucial role, encompassing the incorporated features (as shown in `D'). For low IAA scores (e.g., less than 0.5), they resolve the discord between annotators. Modified data is stored in JSON files via `Save Changes'.
\para{Analysis of annotated data}
As previously noted, we distributed 200 documents among two legal experts for annotation. An analysis of this annotated data reveals the following key points:
\begin{itemize}
     \item Multiple tags have been utilized for labeling a span of text, allowing for more nuanced annotations.
    \item The annotated spans often cross sentence boundaries, indicating the necessity of capturing information that transcends individual sentences.
    \item The length of the annotated spans varies depending on the specific tags. For instance, conveying the essence of a \textit{Homicide\_murder} tag may require only a single sentence, whereas the \textit{Expert\_witness\_testimony} tag often necessitates 4-5 sentences to fully encapsulate the relevant information.
    \item After the completion of the annotation process, another crucial phase comes into play. At this stage, a senior legal expert is tasked with computing the Inter-Annotator Agreement (IAA). It posed a considerable challenge to devise a method that would yield an accurate IAA measure for each annotated document. The complexity arises from the fact that the same text span can be annotated in various sections of the documents, and annotators may use a single tag for a span, while others might employ multiple tags. To address these complexities in the IAA computation, we focused on three key components: i) the annotated span, ii) the tags used to label that span, and iii) the sentence number of the span within the corresponding document. These factors were instrumental in accounting for the subtleties of the annotation process. Subsequently, the super annotators, who are senior legal experts, addressed cases with low IAA scores for each document. This comprehensive approach allowed us to construct our dataset.
\end{itemize}

\begin{figure}[t]
\center
    \caption{An illustration of an annotated document is presented here. In this example, annotators have labeled a span of text with the tags \texttt{Murder\_on\_parole} and \texttt{Second\_murder}, providing a representation of the topical view of the case proceedings.}
    \includegraphics[width=.90\textwidth]{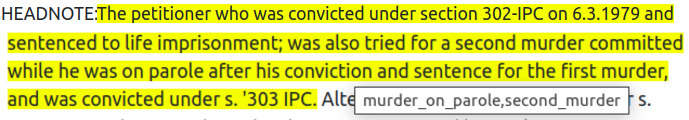}
    
    \label{fig:case_study}
\end{figure}
\subsection*{A Case Study}
To showcase the efficiency of our annotation tool and the significance of the tag list in dataset construction, we present a visual illustration in Figure~\ref{fig:case_study}. In particular, annotators have selected a text span, utilizing the predefined tag list. This case\footnote{\url{http://www.liiofindia.org/in/cases/cen/INSC/1991/208.html}} concerns the conviction of an accused under Section 302 Indian Penal Code for the murder of an individual. Subsequently, during a parole period, the same accused was involved in a second murder. To capture this fine-grained information, they labeled that span of text as \texttt{Murder\_on\_parole} and \texttt{Second\_murder}. Hence, we focus on capturing the thematic representation of the case proceedings by labeling the topics using this tag list (as shown in Table~\ref{tab:tag_list}).

\section{Conclusion and Future Work}
We anticipate leveraging this meticulously annotated dataset in downstream tasks such as prior case retrieval, judgment prediction, etc. Thus, LeDA can be employed to annotate diverse legal documents, taking full advantage of its advanced capabilities. In our future endeavors, we aim to continuously enhance the user interface, incorporating new features suggested by our users. While we currently emphasize murder-related case proceedings, we have plans to expand our focus into other legal domains. To mitigate the time and cost associated with manual annotation, we have planned to explore automated concept extraction using this annotated dataset.

\bibliography{ref}
\bibliographystyle{vancouver}

\end{document}